\documentclass[11pt]{article}

\usepackage[]{acl} % final (camera-ready) mode: author names visible
\usepackage{times}
\usepackage{latexsym}
\usepackage[T1]{fontenc}
\usepackage[utf8]{inputenc}
\usepackage{microtype}
\usepackage{booktabs}
\usepackage{graphicx}
\usepackage{amsmath}
\usepackage{array}
\usepackage{makecell}

\graphicspath{{figures/}}

\DeclareCaptionLabelSeparator{spacedhyphen}{ - }
\newcolumntype{L}[1]{>{\raggedright\arraybackslash}p{#1}}

\makeatletter
\g@addto@macro{\UrlBreaks}{\do\-}
\makeatother

\title{SlopShape: Identifying AI-Generated Commercial Web Content}

\author{Jochen Madler \\
  Sitefire \\
  \texttt{jochen@sitefire.ai}}

\begin{document}

\maketitle

\begin{abstract}
Word-level detectors identify unedited AI-generated text almost perfectly, but the literature documents their brittleness under rewording, and a word-level score neither characterizes a text nor identifies which AI model wrote it. We ask whether AI-generated text can be identified one level deeper, from structural signatures: how information is presented, in what order, with what evidence, and in what voice. We replicate StoryScope \citep{russell2026storyscope}, which showed such patterns for AI-generated fiction, on commercial content: 2,250 pre-ChatGPT human blog posts from 268 company domains against 11,250 AI mirrors from five frontier models. A 203-feature instrument, applied by an LLM and validated in a human gold-annotation session (human-human kappa 0.939, human-model 0.951), detects AI posts from its 176 structural features alone at 97.0 macro-F1 on held-out companies, nearly unchanged (96.1) when every AI post is reworded by its own model. The signal characterizes and attributes: AI posts share a tidy, self-announcing shape, 68.6\% are attributed to the correct source against a 16.7\% chance rate, and human posts occupy rare structural configurations. All effects replicate StoryScope's, consistent in direction and at least as large in magnitude. We release pipeline, instrument, prompts, code, and aggregate artifacts.
\end{abstract}

\section{Introduction}

In commercial search engines like Google or Bing, text carries direct economic value: whether a web page ranks in the top results is directly tied to brand visibility and web traffic. With the rise of AI search engines like ChatGPT and Google AI Mode, web pages are no longer subject to a search engine's algorithm alone, but are also assessed by an LLM. This additional step has changed the rules of Search Engine Optimization (SEO) and sparked a successor practice, Generative Engine Optimization (GEO): editing web page content can lift a page's visibility in AI search answers by up to 40 percent, while traditional tactics like keyword stuffing yield little to no improvement \citep{aggarwal2024geo}, and AI search engines cite systematically different pages than traditional rankings while reducing clicks to the sources they summarize \citep{chen2025dominate,khosravi2026aioverviews}. Section 2 examines the literature in more detail.

At the same time, LLM adoption has driven the marginal cost of producing content toward zero, and low-quality AI-generated text is now so common that it has its own term, ``slop'' \citep{shaib2025slop}. Academic studies of professional content place the share of LLM-assisted text between 9 and 24 percent: about 9 percent of US newspaper articles and up to 24 percent of corporate press releases \citep{liang2025adoption}. Industry analyses report even higher numbers: Graphite \citeyearpar{graphite2025ai,graphite2026ai} estimates that roughly half of newly published English web articles are primarily AI-written, and Ahrefs \citeyearpar{ahrefs2025ai} finds some AI involvement in about three quarters of newly crawled pages. The share of AI-generated text is highest in the commercial web, where pages are most directly coupled to traffic and business incentives. We therefore study AI-generated text where it is most prevalent: commercial web pages, in particular company blog posts.

For identifying AI-generated content, word-level detection works well: on our corpus, a fine-tuned encoder distinguishes unedited AI-generated posts from human writing almost perfectly at near-zero cost (Section 5). Word-level detection has one weakness, however. Changing words is easy, many free tools automate it, and every word-level detector family has a documented failure mode under such adversarial rewording \citep{krishna2023paraphrasing,weberwulff2023testing}. A commercial ecosystem of ``humanizer'' tools has industrialized exactly this attack \citep{masrour2025damage}. So we look one level deeper, for patterns that are much harder to rewrite. We test this directly: rewording every AI post in the test split leaves structural detection nearly unchanged (Section 5.4).

Because word-level detection of reworded AI-generated text is unreliable, policies of platforms like Google have adapted. Google's policy on scaled content abuse targets outcomes rather than methods, sanctioning mass-produced pages ``generated for the primary purpose of manipulating search rankings and not helping users'' whether ``automation, humans or a combination are involved'' \citep{google2024spam}. And instead of words, YouTube uses structural signals like account coordination, upload pacing, and templated narrative patterns \citep{mathur2026slop}.

In this paper, we focus on these structural patterns the platforms observe, and ask whether AI-generated commercial content has a structure, too. And if it does, what that structure looks like: which points AI-generated text makes, in what order, with what evidence, and in what voice - independent of the words being used.

This question has already been answered once, for fiction. StoryScope \citep{russell2026storyscope} showed that AI-generated fiction can be told apart from human fiction by narrative structure alone (93.2 macro-F1 without stylistic signals), with humans occupying statistically rarer regions of narrative space. But fiction is not where AI adoption and economic incentives concentrate. We therefore carry their question, alongside the validated pipeline and methodology, to a domain where text is tied to revenue: commercial blog content. Our claims are focused on detection of single-pass AI-generated blog posts from the five studied AI models, both as generated and after rewording (Section 9).

Figure 1a gives an overview of the study design, and Figure 1b previews the central result. The core methodology of the original paper is held fixed: the same five-stage pipeline (templates, cross-source comparison, feature discovery, deduplication, feature application), the same five AI models, the same classifier protocol and metrics. Our methodology is adapted where the domain demands it, and each adaptation is declared in a deviation register (Section 4.9). Our contributions are:

\begin{itemize}
\item \textbf{The shape of AI-generated commercial writing, and a validated instrument that reveals it.} AI-generated posts share a measurable structural signature, and nine features carry most of the signal (Section 6). The instrument is an 11-dimension commercial template schema discovered bottom-up from human posts, carrying 203 validated features (Section 4). Because the pipeline is LLM-run end to end, we audit it: repeatability runs, a reliability filter applied before any human/AI labels were seen, and a human gold-annotation session that exceeds the original's agreement numbers (Section 7).
\item \textbf{Attribution and characterization.} The shape identifies its author: 68.6 percent of posts are attributed to the correct one of the six sources, against a 16.7 percent chance rate. The shape is also shared: all five AI models crowd into the same common structural configurations, while human posts occupy disproportionately rare ones (Sections 5-6).
\item \textbf{Detection performance, via replication.} Structural features alone detect AI-generated commercial blog posts at 97.0 macro-F1 on held-out data from companies never seen in training, consistent with the original study and larger in magnitude (Section 5). The signal is nearly unchanged when every AI post is reworded by its own model (Section 5.4).
\item \textbf{A verifiable release.} Our data acquisition pipeline, the full instrument, all prompts, code, and the aggregate artifacts behind every exhibit are public, with post-level data available to researchers on request.
\end{itemize}

\begin{figure*}[t]
\centering
\includegraphics[width=0.98\textwidth]{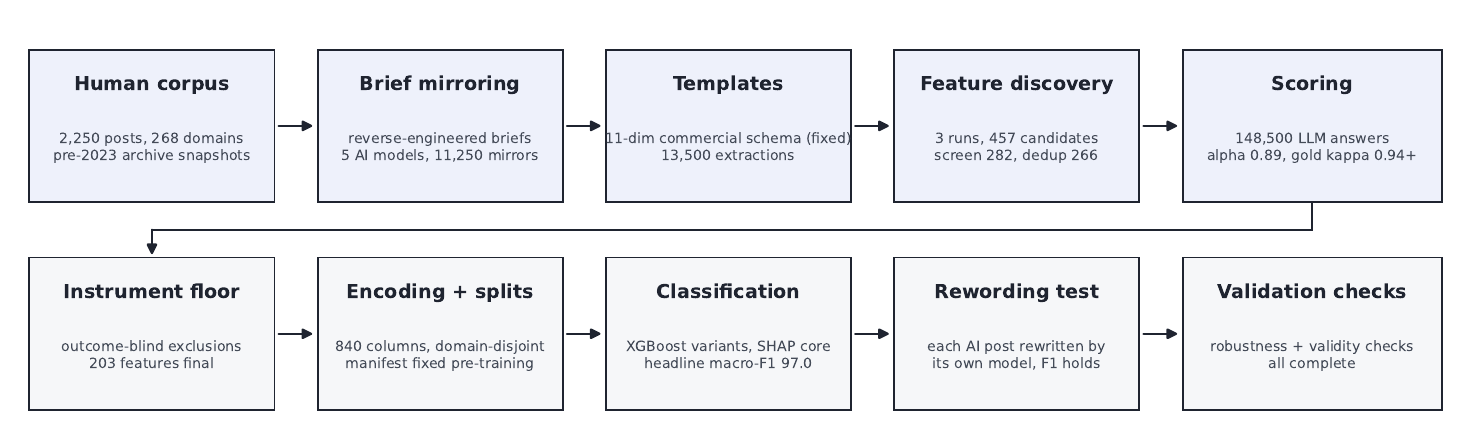}\\[1pt]
{\small (a)}\\[6pt]
\includegraphics[width=0.62\textwidth]{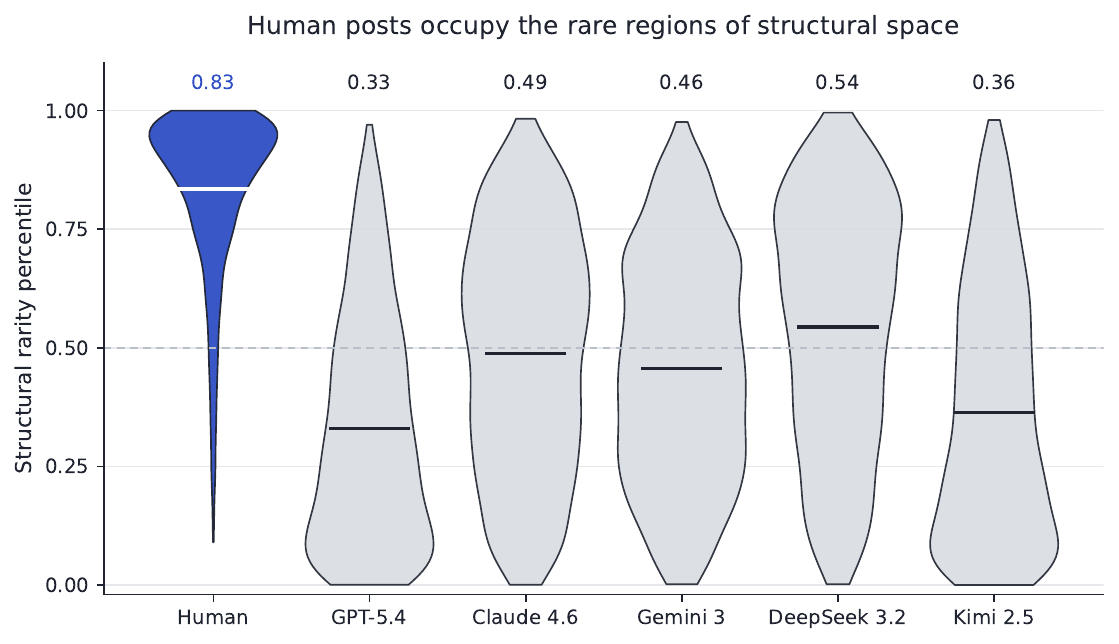}\\[1pt]
{\small (b)}
\caption{(a) Study pipeline, from corpus construction through template extraction, feature discovery, and scoring to classification, the rewording test, and validation. (b) The central result previewed: structural rarity percentile by source. Human posts concentrate in the rarest structural configurations (mean 0.83 vs pooled AI 0.44, Cohen's d = 1.80), while the five AI models crowd the common ones. Section 6 develops this result.}
\label{fig:one}
\end{figure*}

\section{Related Work}

\textbf{AI search and the economics of AI-generated content.} AI search engines answer user prompts with LLM-generated text grounded on retrieved web pages. \citet{aggarwal2024geo} define the term GEO and show that modifying web page content can lift visibility in AI answers by up to 40 percent while traditional keyword stuffing does not. AI search engines cite systematically different sources than classic rankings, with top-5 overlap of only 15-33\% \citep{chen2025dominate}, and getting cited correctly has itself become an optimization target \citep{tian2026citation}. On the user side, AI-generated summaries reduce clicks on the pages they cite: exposure to Google's AI Overviews reduced traffic to certain Wikipedia articles by about 15 percent \citep{khosravi2026aioverviews}. On the producer side, AI search measurably shifts what content communities publish \citep{zhang2026aisearch}.

\textbf{Adoption measurement.} Academic studies already find a sizeable share of AI-generated text in professional writing: about 9 percent of new US newspaper articles \citep{russell2026newspapers}, up to 24 percent of corporate press releases \citep{liang2025adoption}, and more than 5 percent of new English Wikipedia articles \citep{brooks2024wikipedia}. At web scale, a large share of multi-language page translations is already mass machine translation, skewed toward low-quality SEO content \citep{thompson2024machinetranslated}. Industry measurements report even higher shares (Section 1).

\textbf{AI-text detection.} Word-level detectors are the first choice for AI-text detection, and on unedited text they are close to perfect. They span fine-tuned encoders (ModernBERT; \citealp{warner2024modernbert}), stylometric classifiers, TF-IDF baselines, and zero-shot likelihood methods (DetectGPT, \citealp{mitchell2023detectgpt}; Binoculars, \citealp{hans2024binoculars}). Their weakness is rewording, which is easy and widely automated. All four families have documented failure modes under adversarial rewording \citep{sadasivan2023reliably}. Likelihood-based zero-shot detectors collapse under paraphrasing: DetectGPT falls from 70.3 percent to 4.6 percent detection at 1 percent false positives \citep{krishna2023paraphrasing}. Trained classifiers degrade on unseen models, domains, and edits, with accuracy on AI-paraphrased text as low as 26 percent \citep{weberwulff2023testing}. Lexical-tell estimators are defeated by avoiding the tell words \citep{kobak2025vocab}. Provider-side watermarks like the one Anthropic introduced in August 2026 survive light editing but not a complete rewording, by Anthropic's own account \citep{anthropic2026watermark}. The rewording attacks are heterogeneous and hard to anticipate \citep{dugan2024raid}, and there is a commercial tool ecosystem that exploits them \citep{masrour2025damage,perkins2024genai}. The most rewriting-robust result comes from the study this paper replicates: StoryScope's narrative features lost only 1.6 F1 under span-level style rewriting, using the editing protocol of \citet{chakrabarty2025lamp}. We reproduce their word-level baseline suite on our data (Section 5) and run their rewording test on commercial text (Section 5.4).

\textbf{Structural analysis of AI writing, with LLMs as annotators.} StoryScope \citep{russell2026storyscope} built on the NarraBench taxonomy of narrative dimensions \citep{hamilton2025narrabench} to measure discourse-level differences between human and AI fiction, and found structural detectability, model-specific patterns, and a rarity gap. LLM-prose idiosyncrasies have also been formalized from the editing side (the LAMP taxonomy; \citealp{chakrabarty2025lamp}) and the judging side \citep{shaib2025slop}. This paper is a domain-transfer replication of StoryScope, and every deviation from the original is disclosed in a register (Section 4.9). Our pipeline uses LLMs for screening, template extraction, feature discovery, and scoring. We follow the original paper's gold-annotation protocol and extend it with a reliability filter that never uses the human/AI labels (Section 4.5) and a human gold-annotation session (Section 7).

\section{Data}

Our human writing corpus comprises 2,250 human commercial blog posts from 268 company domains, captured in pre-ChatGPT Wayback Machine snapshots dated between 2008 and 2022. The posts come from company websites that sell to other businesses (B2B) and for which organic search is a plausible acquisition channel. Table 1 describes the collection pipeline: we assembled a sampling frame from four public company lists, screened the companies for fit, kept domains with enough articles archived before ChatGPT's release, spot-checked their genre, applied industry quotas, and fetched the qualifying articles through the content filters (600-2,500 words, English, informational genre, near-duplicate removal). Of the 306 selected domains, 264 yielded usable posts (40 had no fetchable qualifying articles, 2 were emptied by the content filters), and 4 replacement domains drawn from the same qualification pool during the fetch gave the 268 corpus domains (see Appendix A). Following the original study's terminology, we define a ``prompt'' as one human blog post together with its brief and the five AI-generated mirror posts derived from it. A seeded, stratified set of 100 prompts, spanning 82 of the 268 domains, is set aside for feature discovery (Section 4.3) and excluded from all classifier splits. We call its 600 posts the feature-discovery set. The remaining 2,150 prompts form the classification corpus, split 198/32/38 by company domain into train/val/test (Section 4.7).

\begin{table*}[t]
\small
\centering
\begin{tabular}{L{4.0cm}L{8.1cm}L{2.4cm}}
\toprule
Step & Rule & Count \\
\midrule
1. Assemble the sampling frame & Sampling frame assembled from four archived public company lists: Inc5000 9,979; YC directory 3,448; FT1000 1,129; G2 519 (with an anti-persona prefilter) & 15,075 domains \\
2. Company-fit screen & 2 LLM judges assess the company fit from firmographics and homepage metadata & 94.3\% keep \\
3. Archive-volume check & At least 25 article URLs archived before 2022-11-30 in the Wayback Machine's snapshot index & 698 qualified \\
4. Genre spot-check & 5 posts/domain checked for English language, informational type, and web page length & 307 keep-eligible \\
5. Industry quotas & Company industry quotas applied (software cap 40\%) & 306 domains \\
6. Fetch and filter & Content filters: 600-2,500 words, English, informational genre, near-duplicate removal & 2,250 posts from 268 domains \\
\bottomrule
\end{tabular}
\caption{Data collection pipeline}
\label{tab:collection}
\end{table*}

The corpus spans seven industry verticals from software/SaaS (26.5\%) to edtech (0.6\%), with the full composition by vertical, source frame, and snapshot year in Appendix A (Table A1). Because we collected as much data as possible from just before ChatGPT's release, the corpus skews recent: 75.5\% of post snapshots are from 2020-2022. We address this with a publication-year check (Appendix G) and disclose it as a limitation (Section 9).

Following the original study's reverse-engineering design, a content brief is inferred from each human post. Like the original's brief prompt, which asks for key characters or settings by name, ours names the publishing company and what it sells, so the publisher commissions the post rather than being its subject; where a post does not name its publisher, the brief says ``the publisher''. Analysis-only labels, such as the post's commercial goal, are never passed to the writer. Each brief is then given to the original's five AI models: gpt-5.4, claude-sonnet-4.6, gemini-3-flash, deepseek-v3.2, and kimi-k2.5, yielding 11,250 AI-generated mirror blog posts. Together, the human posts and their mirrors form a paired corpus: for every human post there are five AI-generated versions written from the same brief. AI mirrors turn out to be slightly longer than their human counterparts (per-model means of 1,064 to 1,541 words, length statistics in Appendix A). To rule out effects driven by post length, we ran a length check (Appendix G) as well as a length-only classifier: given only word count, it cannot separate the classes (Section 5).

In the template-extraction, feature-discovery, and scoring prompts, the web page identity is anonymized. Schema discovery, the stage that derives the 11 description dimensions (Section 4.1), saw only human posts. Feature discovery, the stage that derived the 457 candidate features within those 11 dimensions, saw posts from all six sources without knowing which was which. Nevertheless, a human post could be recognized for a trivial reason: it might be part of an AI model's training data. To rule this out, we ran a memorization check using the 13-gram overlap rule of \citet{brown2020gpt3}: if a human post and its mirror share even one identical 13-word sequence, the pair is flagged as potentially contaminated. Only 0.19\% of pairs were flagged (against 0.0\% for a shuffled-human control), no pair was near-verbatim, and excluding all flagged prompts leaves the headline unchanged in substance (97.2). As a final check, we scanned all posts for mentions of entities that did not yet exist at the post's claimed publication date. Only 0.08\% of posts contain such a mention, and excluding them leaves the headline numbers unchanged.

\section{Measuring Structure}

The measurement pipeline follows the original's five stages, preceded by one new schema-discovery step. Figure 1a gives the overview.

\subsection{Commercial template schema}

Since the original's template schema is fiction-shaped (plot, characters, emotion trajectories), we derived a similar schema for commercial content bottom-up: three independent runs of the same LLM (gpt-5.6-terra) each examined 18 stratified human posts and converged on near-identical dimension systems. The consolidated result is a new 11-dimension schema: purpose, audience, structure and flow, explanation, evidence, voices, actionability, commercial integration, timeliness, page format, and writing style. We kept a dedicated Writing-Style dimension so that style is measured on its own and can be excluded. Every structural result in this paper is computed with these writing style features stripped out, which is what lets us attribute the detection signal to structure rather than wording. In Appendix B, we map the NarraBench fiction dimensions to our commercial ones.

\subsection{Template extraction}

For each of the 13,500 posts - the 2,250 human posts and their 11,250 AI mirrors, six posts per prompt - we extracted one JSON template that describes the post along all 11 dimensions of the schema.

\subsection{Comparison, feature discovery, quality gate, deduplication}

Using the extracted templates, we ran the original study's cross-source comparison stage over the feature-discovery set: the 100 prompts (600 posts) set aside for this purpose and excluded from all classifier training and testing, so that features are never discovered on the same posts they are later evaluated on. The feature-discovery set shares some company domains with the test split, and a dedicated check covers this overlap (Appendix G). For each of the 100 prompts, gpt-5.6-terra received all six templates - the human post's and its five AI mirrors' - side by side and noted where the sources diverge. Three independent LLM discovery runs, each with one specialized prompt per dimension, then turned these divergence notes into 457 features, versus the original's 408. We then applied a quality gate to score each feature on whether it can be answered from the text alone, with locatable evidence, in self-explanatory wording, and without compound constructs. The gate rejected 38.3\% of candidates (see Appendix C), leaving us with 282 features. Finally, we removed duplicate features by clustering their embeddings (F2LLM-4B, single-linkage clustering at cosine 0.85) and keeping only one feature per cluster, which left 266 features. In this embedding step, only 5.7\% of our features were merged versus the original's 25.5\%, possibly because of our quality gate. We include a sweep on the deduplication cutoff threshold that shows the results do not depend on this difference (Appendix G).

\subsection{Feature application}

To turn features into measurements, every blog post is scored against every feature. We used gemini-3.6-flash, one dimension per call, with the model forced to pick exactly one answer from each feature's predefined options. Human posts are scored as extracted from the archived pages and normalized to plain text, which keeps the page title for some posts and drops it for others. AI posts are scored as the models produced them: each opens with a title line, and most carry markdown formatting such as headings, bold text, or lists. Section 4.5 removes the features that respond to this difference. Before committing to the full corpus, we ran two checks on a small sample. First, scoring one dimension at a time answered 99.97\% of items, versus 99.75\% for scoring the whole post in a single call. Small as the difference is, we kept the per-dimension mode (Appendix C). Second, we scored 60 of the posts five times, and agreement across the runs reached a Krippendorff alpha of 0.889 against our 0.8 bar (Section 7). We then scored all 13,500 posts against the 266 features. This took 148,500 scoring calls (13,500 posts x 11 dimensions).

\subsection{Feature reliability filter}

An LLM judge can produce unreliable answers: features whose answer values barely vary, that drift outside the predefined options, or that randomly change from run to run. To remove such features from further analysis, we ran a reliability filter that judges a feature by its result statistics. It excluded 52 features: 11 were degenerate (at least 98\% of posts received the same answer), 3 produced answers outside their predefined options too often (above 2\%), and 38 were unstable across the five repeat scoring runs.

Because human and AI posts reach the scorer in different formats (Section 4.4), a second filter removes features that answer to the format rather than to the post. We rescored three samples from the training and validation splits with only the format changed: 250 AI posts normalized like the human posts (markdown stripped, title kept as a plain first line), the same posts without the title line, and 250 human posts with their page title added. A feature is excluded if its answer distribution moves by more than twice its run-to-run noise plus 0.05 (total variation distance) in any of the three samples; the rule was fixed before the samples were scored. It excluded 11 structural features and no style feature, all tied to titles, headings, or lists: where the payoff and the problem first appear, brand in the title, author identity treatment, heading depth, density, and workflow, bulleted and numbered lists, numbered steps, and numbered main sections. The final instrument therefore has 203 features (versus the original's 304): 176 structural ones (the original's ``narrative-strict'' analogue) and 27 style. Appendix C (Table C1) gives the composition by dimension and answer type.

\subsection{Feature space, variants, encoding}

Our encoding follows the original paper's methodology: one-hot for nominal and binary types, multi-hot for multi-select, integer position encoding for ordinal and scale, NaN for missing (native to XGBoost; \citealp{chen2016xgboost}). The result is a matrix of 12,900 posts x 840 columns: each of the 2,150 classification prompts contributes six posts, and 600 posts from the initial feature-discovery set are excluded. As in the original, we train classifiers on subsets of the instrument: 176 structural features only, style features only (27), all features (203), the nine core features (Section 6), and the nine core features plus model-specific features (30 in total, Section 6).

\subsection{Experimental protocol}

Before training any classifier, we split the corpus by company domain: 1,566 / 294 / 290 prompts (198/32/38 domains), so no domain appears in more than one split. This is stricter than the original's random prompt split (deviation D6) and prevents a company's tone of voice from leaking across splits. Per the original's protocol, we grid-searched hyperparameters on the validation split and trained the final models on training and validation data combined (see Appendix D). Our results are, unless stated otherwise, measured only on the held-out test split: macro-F1 and AUPRC for the human-vs-AI task (Section 5.1 explains this choice), and macro-F1 and accuracy for source identification, the task of predicting whether a post was written by a human or, if not, which of the five AI models wrote it. Confidence intervals come from 10,000 bootstrap resamples. Because posts from the same company resemble each other, the primary intervals resample whole company domains rather than individual prompts.

\textbf{Hypotheses.} We test three hypotheses. The first: style features outperform structural features, the opposite of the original study's finding. The rationale: commercial writing is persuasion in a brand voice, so style could plausibly carry more of the signal than it does in fiction. Our data rejects this hypothesis decisively (Section 5.3). The second: structural detection is robust against rewording, because the features capture what a post says rather than how it formulates it. It holds (Section 5.4). The third: the gold-annotation session must reach human-human and human-model agreement of at least kappa 0.60. Both bars were cleared at 0.939 and 0.951 (Section 7).

\subsection{Rewording test}

To test the second hypothesis, every AI post in the test split is rewritten by the same AI model that generated it. The rewriting follows the editing protocol of \citet{chakrabarty2025lamp}: the model edits its own post span by span, targeting the seven categories of AI-writing artifacts their study identified with professional editors, while keeping every claim, fact, and link. We ensured preservation of content after rewording by running a judge on all 1,450 original-rewritten pairs. In the final pairs, 98.2\% of blog posts preserve all claims and the rest differ in single details. The rewriting attack is substantial: on average, 73\% of a post's 13-word sequences no longer appear verbatim, and in a seeded spot check of 20 posts, 16.0\% of all feature answers change, against 10.9\% between two scorings of the same text (Section~7).

\subsection{Declared deviations}

Table 2 lists the deviations from the original's method that most affect how the results should be read. The full register with all 16 deviations and their defenses is in Appendix E.

\begin{table*}[t]
\small
\centering
\begin{tabular}{L{0.6cm}L{8.6cm}L{5.6cm}}
\toprule
\# & Deviation from the original & Defense \\
\midrule
D1 & Domain: B2B blog posts & The research question \\
D6 & Domain-disjoint splits, single-unblinding holdout & Stricter than their random prompt split \\
D7 & Rarity metric re-implemented from text & Verified on their data \\
D10 & Corpus \textasciitilde 1/4.5 of theirs, feature-discovery set 4.4\% vs \textasciitilde 1\% & Budget, power note in Section 9 \\
D11 & Commercial-native schema discovered from human posts & Fiction lens misfits the domain \\
D13 & Evaluation-stage models upgraded within lineage & The five AI models unchanged \\
D14 & Rewording test: symmetric self-rewrite by all five models & Removes the original's single-model asymmetry \\
\bottomrule
\end{tabular}
\caption{Deviation register (condensed to the most consequential rows).}
\label{tab:deviations}
\end{table*}

\section{Detection Results}

\subsection{Binary detection}

We report macro-F1, an accuracy measure balanced across two classes: it averages the detection quality on human-written and on AI-generated posts equally. Naive accuracy would be misleading, because five of every six posts are AI-generated: a detector that answers ``AI'' for everything already reaches 83.3\% accuracy without catching a single human post.

\begin{table*}[t]
\small
\centering
\begin{tabular}{L{4.8cm}rL{4.5cm}L{1.5cm}L{2.3cm}}
\toprule
Variant / baseline & n feat & Test macro-F1 & AUPRC & Original \\
\midrule
Structural (headline) & 176 & 97.0 (cluster CI 95.5-98.2, prompt CI 95.7-98.1) & 1.000 & 93.2 / .959 \\
Style-only & 27 & 88.1 & .994 & 85.8 / .867 \\
All features (combined) & 203 & 98.0 & 1.000 & 96.0 / .982 \\
Core-only & 9 & 90.1 & .997 & 30 feat: 84.8 / .828 \\
Core + model-specific & 30 & 93.7 & .998 & 101 feat: 91.1 / .934 \\
ModernBERT-base fine-tune & - & 100.0 & 1.000 & 99.9 / 1.00 \\
Stylometric + XGB (144-dim, their spec) & 144 & 100.0 & 1.000 & 99.8 / .999 \\
TF-IDF (1-2g, their 5,000-feature spec) + XGB & 5,000 & 99.3 & 1.000 & 99.7 / .999 \\
\bottomrule
\end{tabular}
\caption{Binary human-vs-AI detection on the held-out test split (macro-F1, original's Table 2 analogue).}
\label{tab:binary}
\end{table*}

The first question is how well structural features alone can distinguish AI-generated from human-written blog posts, and how close they come to the word-level baselines. Table 3 reports classifiers trained on each subset of our instrument as well as the word-level baselines from the original study, all evaluated on the same held-out test data.

\textbf{Result:} We replicate the original study's headline result at larger magnitude. Structural features alone distinguish AI-generated from human-written commercial blog posts at 97.0 macro-F1. The ordering of the feature variants replicates, too: all features (98.0) above structural only (97.0) above style only (88.1), against the original's 96.0 > 93.2 > 85.8.

\textbf{Word-level detection:} the framing of Section 1 holds on this corpus. Unedited single-pass generation is near-perfectly detectable at the word level: the stylometric baseline and ModernBERT both reach a perfect score, mirroring the original's 99.9 and 99.8. Part of the stylometric ceiling comes from formatting rather than wording. Some human posts are interviews, transcripts, or roundups, formats the AI mirrors almost never produce. Restricted to the 86.5\% of human posts that are regular self-contained pages, the structural headline remains 96.2. The zero-shot substitute (a small Qwen2.5-0.5B pair in place of the original's Falcon-7B Binoculars pair) lands at 78.2 where the original's true Binoculars scored 55.9. Moreover, blog post word count by itself cannot separate the classes, consistent with the length check (Appendix G).

As in the original, all feature-based models and the stylometric and TF-IDF baselines are trained on the combined training and validation data. The ModernBERT baseline is an exception: per the original study, we fine-tune it on the training split alone. Two more baselines, a zero-shot detector substitute and a length-only control, are reported in Appendix F, and all values are percentages reported to one decimal. AUPRC measures ranking separation rather than thresholded decisions and is rounded to three decimals (the structural model's stored value is 0.9996), so the structural variants' remaining test errors are threshold errors, not ranking errors. The results are not sensitive to hyperparameter tuning: refitting with the original study's published hyperparameters yields only a 0.4 points difference compared to our tuned configuration on both tasks (Appendix D).

\subsection{Distance to the word-level ceiling}

The best word-level detectors on this corpus are perfect, but the structural model comes within three percentage points: 97.0 macro-F1 against their 100.0 (the original: 93.2 against near-perfect word-level baselines). These word-level detector ceilings are measured on unedited text, and there is extensive literature on the brittleness of word-level detectors against rewording attacks (Section 2). In contrast, our structural feature instrument's accuracy remains nearly unchanged under rewording attacks (Section 5.4).

\subsection{The style-vs-structure direction}

The first hypothesis (Section 4.7) predicted that style features would beat structural features. The data reverses it: the structural model beats the style model by 8.8 macro-F1 points (95\% CI [4.9, 13.2] on the primary company-cluster basis, and [6.5, 11.3] at the prompt level).

Of the 1,740 test posts, the structural model makes 29 errors and the style model 101, and only 9 posts are misclassified by both. If structural features merely re-encoded surface style, the two models should fail on the same posts. Instead, their failure sets are nearly disjoint, which is evidence that the style features and structural features measure two different signals (see Section 8).

\subsection{Detection survives rewording}

The second hypothesis holds. On the test split with every AI post replaced by its reworded version, the structural classifier scores 96.1 macro-F1, against 97.0 on the original posts (95\% CI on the reworded split 94.3-97.6). Of 1,450 reworded AI posts, 18 are misclassified as human-written, against 9 of the same posts before rewording. The result is uniform across the five AI models (95.3-96.5). Style features, which read how a post is worded, lose ground as expected under a wording attack (88.1 to 87.1), and the combined set changes little (98.0 to 97.3). The changed feature answers do not add up to an escape: they are scattered across posts and features rather than directed toward the structural profile typical of human posts, and a classifier that reads the whole 176-feature signature loses less than a point. The original study observed the same for its narrative features in fiction, with a 1.6-point drop on a single-model rewording arm. Ours holds across all five models. The rewriting protocol and verification gates are in Section 4.8 and Appendix J.

\begin{table}[t]
\small
\centering
\begin{tabular}{lrr}
\toprule
Variant & Original & Reworded \\
\midrule
Structural (176) & 97.0 & 96.1 \\
Style-only (27) & 88.1 & 87.1 \\
All features (203) & 98.0 & 97.3 \\
\bottomrule
\end{tabular}
\caption{Detection on the original versus reworded test split (macro-F1).}
\label{tab:reworded}
\end{table}

\subsection{Identifying which AI model wrote a post}

We also trained a model that identifies each post's author among the six sources: the human author or one of the five AI models. On the test split, it assigns 68.6\% of posts to the correct source (macro-F1 68.2, against a 16.7\% chance rate), on par with the original's 68.4\%. Table 5 shows the per-class scores.

\begin{table}[t]
\small
\centering
\begin{tabular}{lrr}
\toprule
Class & Ours & Original (without style) \\
\midrule
human & .944 & .89 \\
gpt & .793 & .73 \\
gemini & .624 & .60 \\
deepseek & .615 & .57 \\
kimi & .579 & .55 \\
claude & .537 & .77 \\
\bottomrule
\end{tabular}
\caption{Per-class F1 for the six-source model (original's Table 11 analogue).}
\label{tab:sixsource}
\end{table}

Human-written posts are identified almost perfectly. The remaining errors are AI posts attributed to the wrong AI model (Figure 2). Posts by Claude Sonnet 4.6 are the hardest to attribute.

\begin{figure}[t]
\centering
\includegraphics[width=\linewidth]{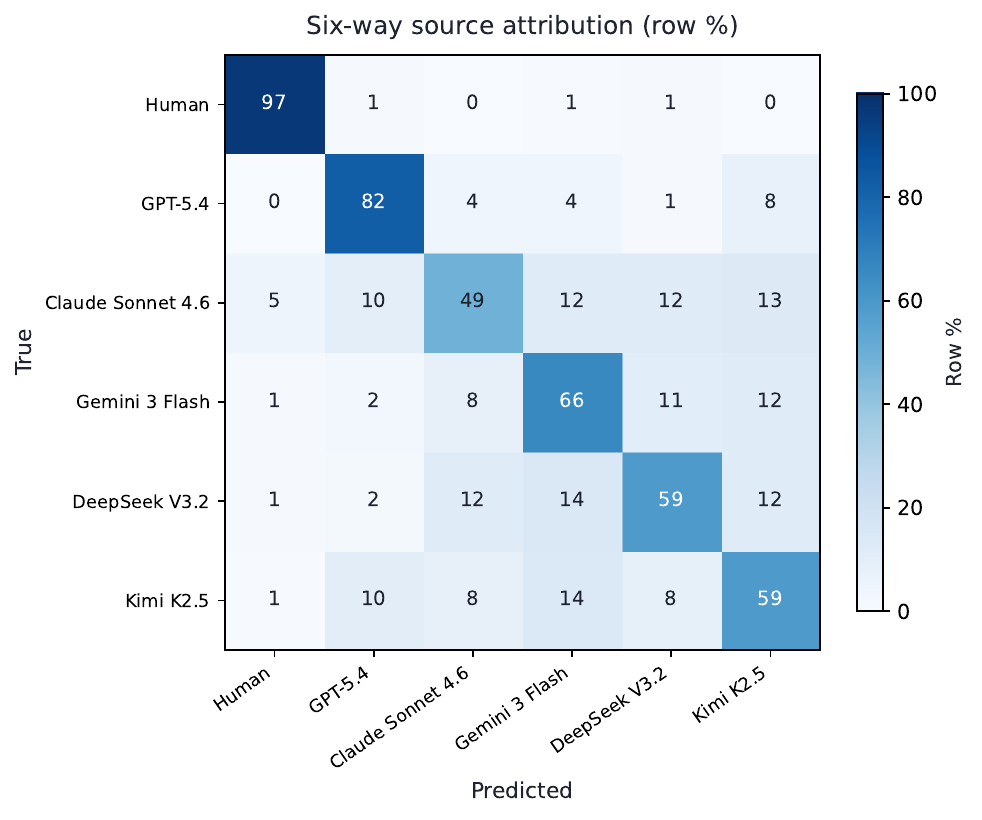}
\caption{Row-normalized test confusion of the six-source model (accuracy 68.6\% vs the 16.7\% chance rate, macro-F1 68.2). Humans are near-perfectly separated. The residual confusion is between AI models, hardest for Claude Sonnet 4.6.}
\label{fig:confusion}
\end{figure}

\section{The Shape of AI Writing}

\begin{figure*}[t]
\centering
\includegraphics[width=0.82\textwidth]{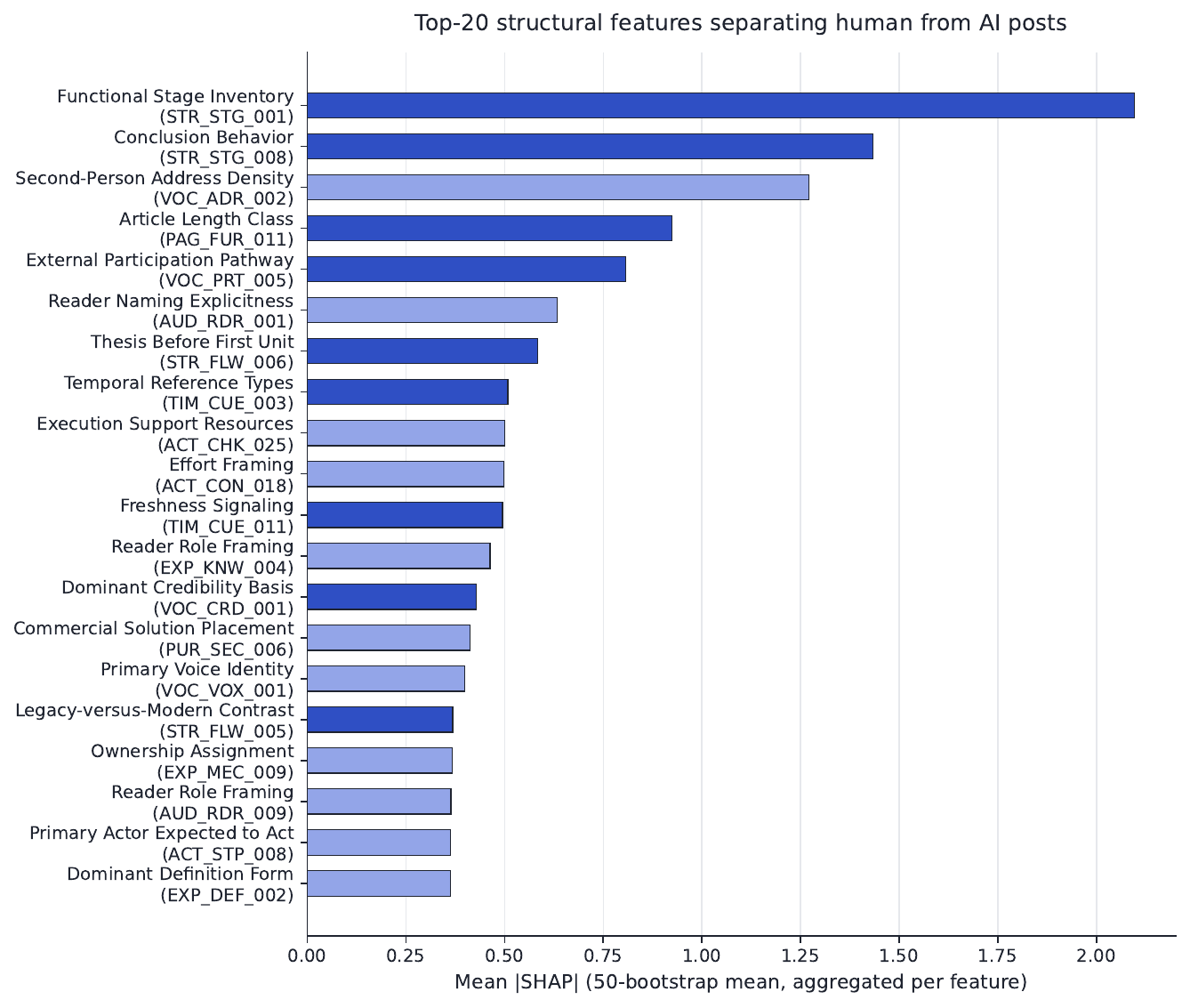}
\caption{Top-20 features of the structural classifier by mean absolute SHAP contribution across bootstrap refits, labeled with their plain-language instrument names (feature IDs in parentheses, full question wording in the released instrument). Dark bars mark the core features (Section 6) that appear among the top 20.}
\label{fig:shap}
\end{figure*}

Figure 3 shows the twenty most important features, ranked by mean absolute SHAP contribution. SHAP (SHapley Additive exPlanations; \citealp{lundberg2017shap}) assigns each feature its share of every prediction the model makes. The most important features describe argument-structure stages, how the reader is addressed, temporal framing, and how sources are disclosed.

Using the original's core-selection procedure, we then distilled this ranking down to a core: the small set of feature values that carry the signal most reliably. A value counts as core if it ranks in the top quartile of SHAP importance, stays important when the model is refit 50 times on resampled data, carries clearly more weight than it would with randomly shuffled labels, and separates human from AI answers by a wide margin that is consistent across all five AI models. This yields 10 core values on 9 distinct features (the original: 33 values on 30 features). These nine features account for 25.6\% of the full model's total feature attribution, and a classifier restricted to them alone still tests at 90.1 macro-F1 (the original: 30 features at 84.8). The small-core phenomenon replicates, with a smaller core and a higher accuracy, potentially because of the quality gate in our feature discovery and because commercial blog posts are more homogeneous than fiction.

\begin{table*}[t]
\small
\centering
\begin{tabular}{L{6.4cm}L{4.4cm}lr}
\toprule
Feature & Value & Leans & Human-AI gap \\
\midrule
Conclusion Behavior (STR\_STG\_008) & restated thesis or reframe & AI & 0.650 \\
Functional Stage Inventory (STR\_STG\_001) & summary or synthesis & AI & 0.609 \\
External Participation Pathway (VOC\_PRT\_005) & absent & AI & 0.592 \\
Legacy-versus-Modern Contrast (STR\_FLW\_005) & no & human & 0.497 \\
Thesis Before First Unit (STR\_FLW\_006) & no & human & 0.419 \\
Freshness Signaling (TIM\_CUE\_011) & modern or next-generation framing & AI & 0.360 \\
Dominant Credibility Basis (VOC\_CRD\_001) & confident institutional explanation & AI & 0.347 \\
Functional Stage Inventory (STR\_STG\_001) & mechanism explanation & AI & 0.285 \\
Temporal Reference Types (TIM\_CUE\_003) & historical date or period & human & 0.283 \\
Article Length Class (PAG\_FUR\_011) & short, under 800 words & human & 0.274 \\
\bottomrule
\end{tabular}
\caption{All core values, with the class each one points to (original's Table 15 analogue).}
\label{tab:core}
\end{table*}

Each core value points toward one author class, and together the directions form a consistent pattern. We interpret only concrete, directly observable features, within what the validity results of Section 7 support. The values most commonly produced by AI models assemble into what we call the tidy, self-announcing blog post: it states its thesis and announces its structure before the first section, presents its subject as the modern answer to a legacy way of working, explains in a confident institutional voice, and closes with a stage that summarizes or restates the thesis. In a typical AI post, the thesis is stated and the flow announced before the first section: ``In this post, we'll cover why onboarding stalls, three fixes that work, and how to measure the difference.'' The subject is set against the old way: ``Paper checklists were built for a slower era.'' And the close restates the thesis: ``In short, structured onboarding saves time.'' The human-leaning values describe posts without those signposts: no thesis up front, no legacy-versus-modern contrast, more references to specific dates and periods, and more short posts. The length-matched check in Section 7 shows the headline result does not depend on post length.

\textbf{Rarity.} Following the original study and the tradition of \citet{torrance1966tests}, we measure statistical rarity as a proxy for originality. Rarity is a per-post statistic: for each post, we compute the mean Euclidean distance to its 25 nearest neighbors in the z-scored structural feature space, and convert it to a percentile against the pooled train+val distribution as reference. A post with a common structure sits close to many neighbors and scores low. A post whose structure few others share sits far from its neighbors and scores high. We compute these statistics over all 12,900 posts, and on the original's test-only basis the picture is the same (rarest 1\%: 36 human vs 0 AI, full tail table in Appendix F). Human posts occupy the rarest regions of this space and AI posts the densest (Table 7 and Figure 1b): the mean human rarity percentile is 0.835 against 0.436 for AI, a gap of Cohen's d = 1.80 (the original: 0.71 vs 0.49, d = 0.83). The contrast is even sharper in the tail. Among the posts whose rarity exceeds the 99th percentile of the reference distribution, 146 are human and 2 are AI, where the original study's rarest 1\% split roughly evenly (42 vs 41). The gap is equally visible in ranking terms: rarity alone separates the classes at AUC 0.898 (original 0.73), and the human post is the rarest of its six-post prompt group in 85.6\% of prompts, against a 16.7\% chance rate (original 57.8\%). Per-model mean rarity orders the AI models deepseek 0.543 > claude 0.489 > gemini 0.456 > kimi 0.363 > gpt 0.330. DeepSeek sits closest to humans, and the length-matched check (Appendix G) shows this proximity is structural rather than length-driven: the per-model ordering is preserved on the length-matched subset.

\begin{table}[t]
\small
\centering
\begin{tabular}{L{3.4cm}L{1.6cm}L{1.6cm}}
\toprule
Statistic & Ours & Original \\
\midrule
Human mean rarity percentile & 0.835 & 0.71 \\
AI mean rarity percentile & 0.436 & 0.49 \\
Cohen's d & 1.80 & 0.83 \\
Rarity AUC & 0.898 & 0.73 \\
Human rarest-of-prompt (chance 16.7\%) & 85.6\% & 57.8\% \\
Rarest decile share (human / AI) & 47.3\% / 3.0\% & 24.7\% / 7.1\% \\
\bottomrule
\end{tabular}
\caption{Rarity statistics (all 12,900 posts, percentile reference train+val).}
\label{tab:rarity}
\end{table}

The rarity gap also has a geometric reading: human posts are spread out widely in structural feature space while AI posts cluster together (Appendix F). From the six-source model's per-class SHAP attributions, we selected 28 model-specific features - features whose importance concentrates on one source (human 17, gemini 4, gpt 3, deepseek 2, kimi 2, claude none; the original: 75 features). A classifier on the nine core features plus these 28 model-specific features (30 distinct features) tests at 93.7 macro-F1, and the per-source feature lists are in Appendix F (Table F3).

\section{Robustness and Instrument Validity}

We ran nine robustness checks, every one under the original's protocol (final models retrained on train+val). The rewording test, the tenth robustness result, attacks the input text rather than the pipeline and is reported with the detection results (Section 5.4). Table 8 summarizes them one line each, and the full table with complete results and caveats is Appendix G.

\begin{table*}[t]
\small
\centering
\begin{tabular}{L{0.5cm}L{4.6cm}L{9.6cm}}
\toprule
\# & Check & One-number result \\
\midrule
1 & Length confound & Headline 96.9 on a length-matched test subsample (vs 97.0 unmatched) \\
2 & Publication-year check & Predicting a human post's publication period from structure is at chance (macro-F1 49.1) \\
3 & Template-vs-direct discovery & Same candidate yield, only \textasciitilde 23\% semantic feature overlap \\
4 & Deduplication cutoff sweep & 200-282 features across cutoffs 0.70-0.95, sensitivity low \\
5 & Feature-discovery-set domain overlap & Headline 96.5 excluding the 11 overlapping domains \\
6 & Memorization & 0.19\% of pairs flagged, filtered headline 97.2 \\
7 & Vertical heterogeneity & No significant variation across verticals (Kruskal-Wallis p = 0.107) \\
8 & Learning curve & 95.4 / 96.4 / 96.4 / 97.0 at 25 / 50 / 75 / 100\% of training data \\
9 & Format sensitivity & Rescoring test posts with markdown stripped, title removed, or title added changes few calls: 249 and 247 of 250 AI posts still detected, 15 vs 18 of 235 human posts called AI \\
\bottomrule
\end{tabular}
\caption{Robustness check summary (full versions in Appendix G).}
\label{tab:robustness}
\end{table*}

Two checks target the design confounds declared in Section 3. On a test subsample in which human and AI posts are matched on length (1,545 posts), detection is unchanged at 96.9 versus 97.0 unmatched (the original: 93.2 to 93.2), the rarity gap survives the matching (d = 1.83), and word count is nearly uncorrelated with rarity. Structural features also do not secretly encode when a post was written: predicting a human post's publication period from them performs at chance (macro-F1 49.1; 187 vs 1,622 posts, where always guessing the larger class scores 47.3), and removing the most year-associated features leaves the headline flat (96.7-97.0).

The remaining checks close the other declared exposures. Excluding the 0.19\% of posts flagged by the memorization check leaves the headline at 97.2, and excluding the 11 company domains that the feature-discovery set shares with the test split leaves it at 96.5. Varying the deduplication cutoff between 0.70 and 0.95 changes the feature count (200 to 282) but not the conclusions. Discovering features directly from raw text instead of from templates yields as many candidates but substantially different ones (about 23\% overlap), so the discovery pathway shapes which features are found - and the template pathway is the one validated end to end. Detection strength does not vary significantly across industry verticals (p = 0.107, per-vertical macro-F1 between 94.8 and 100), consistent with the original's finding of no topic effect. Finally, the learning curve reads 95.4, 96.4, 96.4, and 97.0 at 25, 50, 75, and 100\% of the training data (Appendix G).

\textbf{Repeatability.} We scored the same 60 posts five independent times with the scoring model. Agreement across the five runs is Krippendorff alpha 0.889 (\citealp{krippendorff2004content}; our 0.8 bar; original 0.90), mean pairwise Cohen kappa 0.888 (\citealp{cohen1960kappa}; original 0.89), and pairwise exact agreement 0.891.

\textbf{Human gold validation.} To validate that the LLM scorer measures what a careful human reader would measure, two annotators independently answered 20 features on 12 posts, scored in the encoded representation on the 19 features that belong to the final instrument (228 items each). We fixed the acceptance thresholds before annotation began: human-human and human-model kappa of at least 0.60. Table 9 reports the result next to the original's, with each kappa computed as Cohen's kappa on the pooled item-level contingency over all scored items. Our human-human kappa is 0.939 against their 0.739, and our mean human-model kappa is 0.951 against their 0.839. The LLM scorer agrees with the human annotators more closely than the original's did, and more closely than the original's annotators agreed with each other. Each annotator's individual agreement with the model clears the 0.60 bar as well. In a separate audit, the annotators reviewed the model's calls on which features count as style rather than structure, and endorsed that boundary (all kappas above the 0.75 bar, detail in Appendix H).

\begin{table}[t]
\small
\centering
\begin{tabular}{L{2.2cm}L{1.3cm}L{1.0cm}L{1.7cm}}
\toprule
Comparison & Agreement & Kappa & Original \\
\midrule
Human-human & 94.20\% & 0.939 & 76.85\% / 0.739 \\
Annotator A vs model & 95.09\% & 0.9482 & 91.67\% / 0.9056 \\
Annotator B vs model & 95.54\% & 0.9529 & 79.86\% / 0.7724 \\
Mean human-model & - & 0.951 & 0.839 \\
\bottomrule
\end{tabular}
\caption{Human gold validation (original's Table 7, encoded basis, kappas computed as Cohen's kappa on the pooled item-level contingency).}
\label{tab:gold}
\end{table}

\textbf{Disclosures.} The reliability filter's 52 exclusions and the 11 format-sensitivity exclusions are itemized in Section 4.5, and the answer distributions, drift rates, and per-feature stability records are published in the release. Both annotators are company-affiliated (Section 9).

\section{Discussion and Conclusion}

We asked whether AI-generated text has a shape: a structural signature in how information is presented, in what order, with what evidence, and in what voice, independent of the words being used. Within the studied scope, the answer is yes. Commercial blog posts have no plot, no characters, and no narrative arc, yet the structural signature transfers. Detection performance shows the shape is a real signal and not an artifact of our instrument: structural features alone reach 97.0 macro-F1, and every phenomenon of the original study replicates in the new domain, consistent in direction and at least as large in magnitude.

The core feature values in Table~6 spell out a shape that we call the tidy, self-announcing blog post. AI-generated blog posts state their thesis before the first section, announce its flow, set their subject against a legacy way of working, close with a summary, and speak in a confident institutional voice. The shape also identifies the author: our classifier assigns 68.6\% of posts to the correct one of the six sources, far above the 16.7\% chance rate (Section~5.5). And the distinct shape is shared across AI models: human posts sit in the rare regions of structural space, while the five AI models concentrate in common ones (Section~6). \citet{mathur2026slop} observed at platform scale that mass-produced generative content can be seen as unique variations of functionally identical material. At the level of company blog posts, our measurement confirms this.

The shape also goes deeper than the word level. Rewording attacks replace most of the surface phrasing, but the structural feature detection accuracy remains nearly unchanged (Section~5.4). The error-overlap result points the same way: the structural and style models fail on almost entirely different posts, so the two feature sets carry different signals.

For search engine platforms, the practical reading is that structural features are a candidate signal for detecting AI-generated content that is robust against word-level attacks. Google's policy on scaled content abuse already targets outcomes rather than methods (Section~1), and our structural instrument fits this reasoning. The instrument also reads both ways: the same features that flag the common structural configurations AI posts use also locate the rare configurations of human writing.

There are three limitations to our instrument. First, we cover single-pass generation from the five state-of-the-art AI models. Second, the instrument is LLM-run end to end. Like the original study, our validity rests on the validity of Section~7. Third, we tested the robustness of our instrument only against rewording attacks, not against the myriad of other, potentially more sophisticated humanizer and paraphrase tools. Those attack classes remain an interesting avenue for future research (Section~9).

To facilitate this research, we release the pipeline, methodology and code together with the paper.

\section{Limitations}

\textbf{Scope of the detection claim.} Our design measures single-pass generation, both as generated and after rewording (Section 5.4): every AI post is a model's first-shot response to a reverse-engineered brief, optionally rewritten by the model that generated it - not humanized, heavily post-processed, or human-AI collaborative content.

\textbf{Brief lossiness.} The mirror mechanism is informationally asymmetric: the human author wrote from full business context, while each AI model conditions only on a reverse-engineered brief. Some human-AI structural differences may therefore reflect what the brief fails to carry rather than how models write. We inherit this limitation from the original study.

\textbf{Input format.} Human posts are scored as extracted and normalized, AI posts as generated. The format-sensitivity filter removes the features this difference moves beyond scorer noise, and the test-split check shows the headline classifier is stable to it (Appendix G), but smaller format effects on the remaining features cannot be ruled out.

\textbf{Publication-year confound (inherent).} Our human posts predate ChatGPT and mirrors were generated in August 2026, so publication period and authorship are confounded. The publication-year check provides evidence that structural features cannot predict when a human post was published, and removing the most year-associated features does not change the headline result. Still, same-period human-vs-AI comparison remains open.

\textbf{Magnitude interpretation.} Every effect we report is consistent in direction with the original and at least as large in magnitude. The discovery stage optimizes features to separate these six sources on the corpus, and a commercial-native schema fitted to the domain probably amplifies the effect. We therefore do not claim that commercial content is intrinsically more separable than fiction, only that the underlying effect replicates.

\textbf{Corpus scale and composition.} The corpus has 2,250 prompts against the original's 10,272, and the feature-discovery set is 4.4\% of the corpus versus their roughly 1\%. Our bootstrap CI widths are adjusted accordingly, but per-class attribution and top-tail rarity statistics are underpowered relative to the original. The sampling corpus is software-heavy, US-heavy, and 75.5\% of post snapshots date from 2020-2022.

\textbf{Attack coverage.} Our rewording test covers rewording by the generating models themselves, following an editing protocol of \citet{chakrabarty2025lamp}. It does not cover a range of humanizer or paraphrase tools. Claims about those attack classes remain open.

\textbf{Competing interests.} The author operates Sitefire, a commercial GEO product (formal statement in the Statements section).

\section*{Statements}

\textbf{Data availability.} The release package is at \url{https://github.com/pulse-energy-eu/slopshape}. Publicly released: the sampling ledger with a deterministic fetch pipeline that rebuilds the exact human corpus from public archives, the template schema with its NarraBench mapping, the complete 203-feature instrument, the complete prompt set (including the rewording-attack prompt), the analysis code (including the fork patch against the original's released pipeline and the rewording verification gates), and the aggregate artifacts and regeneration scripts behind every number and figure in this paper. Available to researchers on request, under a non-commercial research agreement: the per-post feature answers, deterministic refits of the trained models, and the generated briefs and mirrors. We gate these and only these because the ready-to-use scoring assets are the direct substrate of a commercial product and because the mirrors embed content derived from copyrighted source posts. We do not redistribute the copyrighted human posts themselves, and everything needed to audit the paper's claims can be rebuilt with the released pipeline.

\textbf{Ethics.} Human posts are publicly published marketing content collected from public web archives and are not redistributed. No personal data beyond published bylines is processed.

\textbf{Competing interests.} The author operates Sitefire, a commercial GEO product. The study's methods, instrument, prompts, code, and aggregate artifacts are publicly released, with post-level data available to researchers on request, so that every claim can be verified independently of that interest.

\textbf{AI disclosure.} Beyond their role in the method, AI models and coding agents (Claude Code) were used to write and test the analysis code, generate tables and figures, and draft and edit the text. The author takes full responsibility for the content.

\textbf{Acknowledgments.} We thank J. Russell for correspondence resolving ambiguities in the original study's reported constants.

\bibliography{references}

\appendix

\makeatletter
\@addtoreset{table}{section}
\@addtoreset{figure}{section}
\makeatother
\renewcommand{\thetable}{\thesection\arabic{table}}
\renewcommand{\thefigure}{\thesection\arabic{figure}}

\section{Data collection detail}
\label{app:a}

This appendix carries the corpus composition, mirror coverage, and funnel mechanics summarized in Section 3.

Of the 306 selected domains, 264 yielded usable posts (40 had no fetchable qualifying articles, 2 were emptied by the content filters), and 4 replacement domains drawn from the same qualification pool during the fetch gave the 268 corpus domains. A per-domain cap of 20 posts was applied (four domains reach at most 29). The largest single domain contributes 1.3\% of the corpus.

\begin{table*}[t]
\small
\centering
\begin{tabular}{L{2.6cm}L{11.6cm}}
\toprule
Axis & Shares (\%) \\
\midrule
Verticals & software/SaaS 26.5, e-commerce 24.3, services 17.4, fintech 15.7, devtools 11.7, health 3.8, edtech 0.6 \\
Source frame & YC 57.2, Inc5000 33.7, G2 6.8, FT1000 2.3 \\
Snapshot years & 2008-17: 8.5, 2018-19: 16.0, 2020-21: 41.9, 2022: 33.6 \\
\bottomrule
\end{tabular}
\caption{Data composition.}
\label{tab:composition}
\end{table*}

\begin{table}[t]
\small
\centering
\setlength{\tabcolsep}{4pt}
\begin{tabular}{lrrr}
\toprule
Source & Posts & Mean words & \makecell[l]{Within 10\% of\\source length} \\
\midrule
human & 2,250 & 1,186 & - \\
gpt-5.4 & 2,250 & 1,541 & .07 \\
deepseek-v3.2 & 2,250 & 1,447 & .16 \\
gemini-3-flash & 2,250 & 1,283 & .41 \\
claude-sonnet-4.6 & 2,250 & 1,279 & .66 \\
kimi-k2.5 & 2,250 & 1,064 & .47 \\
\bottomrule
\end{tabular}
\caption{AI mirror coverage and length. The last column reports, per model, the share of mirrors whose word count lands within 10 percent of their human source's word count.}
\label{tab:mirrors}
\end{table}

\begin{figure}[t]
\centering
\includegraphics[width=\linewidth]{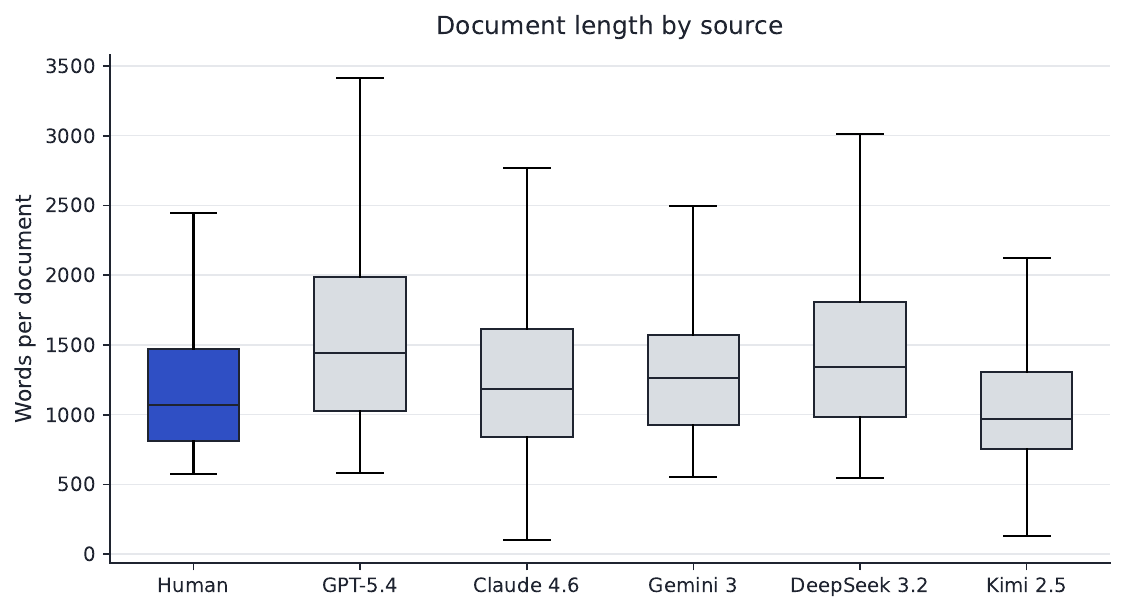}
\caption{Word-count distributions by source (boxes = IQR, whiskers to 1.5 IQR, outliers hidden). Four of the five AI models produce longer posts than their human sources on average, while kimi-k2.5 runs slightly shorter.}
\label{fig:lengths}
\end{figure}

\section{Schema and NarraBench mapping}
\label{app:b}

This appendix maps each NarraBench dimension \citep{hamilton2025narrabench} to its counterpart in the commercial schema of Section 4.1.

\begin{table*}[t]
\small
\centering
\begin{tabular}{L{3.2cm}L{2.0cm}L{9.4cm}}
\toprule
NarraBench dimension & Status & Commercial schema mapping \\
\midrule
Agent & ADAPTED & Voices and Sources: authors, experts, brands, customers, readers as discourse roles \\
Social Network & REPLACED & Voices and Sources: source arrangements and reader interaction, not interpersonal plot relations \\
Event & ADAPTED & Audience/Problem/Stakes and Timeliness: business triggers, policy changes, lifecycle moments \\
Plot & REPLACED & Structure and Flow: movement from problem or context to explanation, recommendation, resolution \\
Structure & ADAPTED & Structure and Flow plus Page Format and Navigation \\
Setting & ADAPTED & Audience/Problem/Stakes: the reader's professional, market, regulatory, technical context \\
Time & KEPT & Timeliness: chronology, temporal reference, change, timing relevance \\
Revelation & REPLACED & Explanation Depth plus Evidence and Proof: staged explanation and substantiation \\
Perspective & ADAPTED & Voices and Sources: institutional, founder, expert, interviewer, direct-reader viewpoints \\
Style & KEPT & Writing Style: the single mandated style bucket \\
\bottomrule
\end{tabular}
\caption{NarraBench mapping (kept / adapted / replaced).}
\label{tab:narrabench}
\end{table*}

\section{Instrument detail}
\label{app:c}

This appendix carries the final instrument's composition, the quality-gate detail behind Section 4.3, and the scoring-mode gate behind Section 4.4.

\begin{table*}[t]
\small
\centering
\begin{tabular}{L{3.4cm}rL{5.2cm}L{4.6cm}}
\toprule
Dimension & Features & Types (categorical/binary/ordinal/scale/multi) & Example feature \\
\midrule
Actionability & 33 & 18/5/5/1/4 & Conditional Logic Presence \\
Writing style & 27 & 14/4/4/4/1 & Dominant Sentence Complexity \\
Audience & 22 & 8/7/3/4/0 & Opening Reader Question Invitation \\
Explanation & 21 & 13/4/2/1/1 & Caveat Presence \\
Timeliness & 20 & 11/1/4/2/2 & Primary Change Story \\
Voices & 20 & 9/4/4/1/2 & Primary Reader Role \\
Structure and flow & 18 & 10/3/3/1/1 & Primary Repeated Building Block \\
Evidence & 17 & 5/6/3/1/2 & Attribution Density \\
Commercial integration & 14 & 5/4/3/2/0 & Proof of Capability Presence \\
Purpose & 8 & 6/2/0/0/0 & Dominant Communicative Purpose \\
Page format & 3 & 3/0/0/0/0 & Article Length Class \\
\bottomrule
\end{tabular}
\caption{Final 203-feature instrument composition.}
\label{tab:instrument}
\end{table*}

\textbf{Quality gate.} The gate never sees human/AI labels, and its dominant rejection defect was non-exhaustive value lists. The gate prompt and all rejection reasons are preserved in the release. The deduplication cutoff sweep is check 4 in Appendix G.

\textbf{Aspect-based application check.} Scoring one dimension at a time versus scoring a whole post in one call, on 12 posts: 99.97\% vs 99.75\% of items answered, cross-mode agreement 0.774 (original: 95.4\% vs 68.4\%, with the single-call mode systematically dropping later dimensions). We kept per-dimension scoring, per the original.

\section{Protocol detail}
\label{app:d}

This appendix records the executed hyperparameter grids and the hyperparameter sensitivity result behind Section 4.7.

The binary task's grid comprised 108 configurations over n\_estimators, max\_depth, reg\_lambda, and scale\_pos\_weight, and the six-source task ran its own 27-configuration grid, both centered on the original's published constants (500/7/1.0). Hyperparameter sensitivity: refitting with those constants, untuned, gives binary 96.6 (AUPRC 0.999) and six-source macro-F1 68.6 (accuracy 69.1) - within 0.4 points of the tuned configurations on both tasks.

\section{Full deviation register}
\label{app:e}

This appendix carries the full deviation register with each deviation's defense.

\begin{table*}[t]
\small
\centering
\begin{tabular}{L{0.6cm}L{8.6cm}L{5.6cm}}
\toprule
\# & Deviation from the original & Defense \\
\midrule
D1 & Domain: B2B blog posts & The research question \\
D2 & Brief extractor gemini-3-flash (their gemini-2.5-flash is deprecated) & Declared \\
D3 & Brief prompt adapted: names the publisher as commissioner, anti-quotation clause added & Parity: the original's brief names characters and settings \\
D4 & Dedup cutoff 0.85 retained from the original (not re-selected) & Parity, sweep published (Appendix G, check 4) \\
D5 & Style audit 3 runs with agreement statistics (theirs: 1 run) & Exceeds the original \\
D6 & Domain-disjoint splits, single-unblinding holdout (theirs: random prompt split) & Stricter \\
D7 & Rarity metric re-implemented from text & Verified on their data, magnitudes disclaimed \\
D8 & Extraction and discovery via OpenAI direct API & Provider plumbing \\
D9 & Encoder fixes (ordinal by taxonomy position, nominal one-hot) & Their released code deviated from their own paper \\
D10 & Corpus scale \textasciitilde 2,250 prompts (\textasciitilde 1/4.5 of theirs), feature-discovery set 4.4\% of corpus vs their \textasciitilde 1\% & Budget and domain scarcity of archived pre-2022 posts, power note in Section 9 \\
D11 & Commercial-native template schema discovered from human posts (theirs: NarraBench fiction schema) & Fiction lens misfits commercial content, mechanism unchanged, mapping table published \\
D12 & Mirror max\_tokens 8,000 (theirs: 128k/65k) & Matched to commercial target lengths \\
D13 & Evaluation-stage models upgraded within family lineage, the five AI models unchanged & Same-variant newer versions for non-generating tasks \\
D14 & Rewording test: each of the five AI models rewrites its own posts, all 1,450 test mirrors (theirs: 278 posts, one model as rewriter) & Realistic production pattern, removes the original's single-model asymmetry, yields a per-model breakdown \\
D15 & Rewording prompt is instruction-only (the original's 25 few-shot professional examples are fiction rewrites) & Domain adaptation, declared - not parity \\
D16 & Content-preservation constraint added (keep claims, facts, links) & Verified by a full claim-preservation census (Appendix J) \\
\bottomrule
\end{tabular}
\caption{Deviation register (full defenses).}
\label{tab:deviationsfull}
\end{table*}

\section{Additional results}
\label{app:f}

This appendix carries the additional detection baselines, the LDA figure, the rarity tail table, the model-specific feature table, and the geometry numbers.

\begin{table*}[t]
\small
\centering
\begin{tabular}{L{4.9cm}L{0.9cm}L{1.7cm}L{3.4cm}L{2.8cm}}
\toprule
Variant / baseline & n feat & Test macro-F1 & AUPRC & Original \\
\midrule
Binoculars-STYLE (small-pair substitute, context only) & - & 78.2 & .980 & (their true Binoculars: 55.9 / .404) \\
Length-only logistic & 1 & 45.5 & .871 (prevalence baseline .833 at the 5:1 ratio) & 55.9 \\
\bottomrule
\end{tabular}
\caption{Additional binary detection baselines (rows not in Table 3).}
\label{tab:additionalbaselines}
\end{table*}

Protocol notes: the feature-variant models and the stylometric, TF-IDF, and length baselines are retrained on train+val per the original's protocol. The ModernBERT row (single seed, their 512-token 3-epoch configuration) and the Binoculars-style row (Qwen2.5-0.5B base/instruct pair in place of their Falcon-7B pair) are single fine-tunes or zero-shot substitutes, disclosed.

\setcounter{figure}{1}
\begin{figure}[t]
\centering
\includegraphics[width=\linewidth]{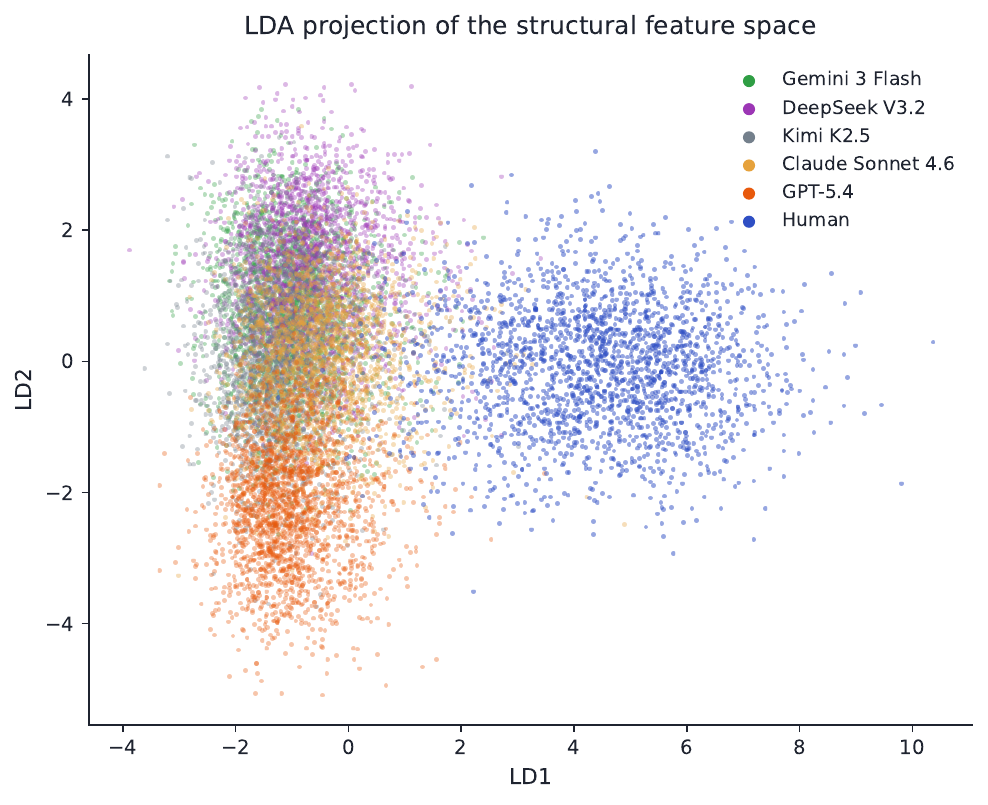}
\caption{Two-component LDA projection of the encoded structural feature space (12,900 posts). Human posts separate along LD1, while the five AI models overlap heavily with each other, consistent with the original's geometry finding.}
\label{fig:lda}
\end{figure}

\begin{table*}[t]
\small
\centering
\begin{tabular}{L{1.6cm}L{2.4cm}L{4.4cm}L{2.6cm}L{3.0cm}}
\toprule
Tail & Human posts & AI posts (five models pooled) & Human share of tail & Original (test basis) \\
\midrule
Rarest 1\% & 146 & 2 & 98.7\% & 42 human vs 41 AI \\
Rarest 5\% & 586 & 85 & 87.3\% & - \\
Rarest 10\% & 1,016 & 319 & 76.1\% & - \\
\bottomrule
\end{tabular}
\caption{Rarity tail composition.}
\label{tab:tail}
\end{table*}

Cohen's d uses the unweighted root-mean-square of the two group SDs (the original pooled by group size).

\begin{table*}[t]
\small
\centering
\begin{tabular}{L{1.4cm}L{13.0cm}}
\toprule
Source & Model-specific features (up to 5) \\
\midrule
human & Execution Support Resources (ACT\_CHK\_025); Link and Resource Placement (ACT\_CHK\_032); False Default Challenged (AUD\_PRB\_005); Legacy Versus Modern Contrast (AUD\_PRB\_006); Reader Situation Mirroring (AUD\_RDR\_006) \\
gpt & Numeric Density (EVD\_NUM\_001); Quantitative Claim Sourcing (EVD\_NUM\_006); Article Length Class (PAG\_FUR\_011) \\
claude & - \\
gemini & Opening Reader Question Invitation (AUD\_HOK\_006); Self-Reference Frequency (COM\_ROL\_005); Legacy-to-Modern Commercial Contrast (COM\_ROL\_011); Evidence Limitation Disclosure (EXP\_KNW\_009) \\
deepseek & Change Requested of Reader (PUR\_JOB\_004); Conditional Routing Density (STR\_UNI\_003) \\
kimi & Product as Stakes Resolution (AUD\_STK\_007); Commercial CTA Intensity (VOC\_PRT\_003) \\
\bottomrule
\end{tabular}
\caption{Model-specific features by source (top features by six-source SHAP concentration). The table lists up to five features per source, with per-source totals in Section 6.}
\label{tab:modelspecific}
\end{table*}

Selection is by per-class SHAP concentration, with the full criteria in the release. The original selected 75 features carrying 90 per-source values (counts 32/26/11/11/7/3).

\textbf{Geometry.} The average distance from a human post to its ten nearest neighbors in the encoded structural feature space is 1.42 times larger than for an AI post, the human class as a whole is 1.41 times more dispersed than the average AI class, and the centers of the human and AI point clouds sit clearly apart (centroid distance 11.3 in z-scored units). The original found the same picture in fiction - humans more dispersed, the AI models clustered together (+22\% dispersion, 1.13x neighbor radius) - and our version of it is again larger.

\section{Robustness checks, full table}
\label{app:g}

This appendix carries the full results and caveats behind the check summary of Table 8. Every check runs under the original's protocol (final models retrained on train+val).

\begin{table*}[t]
\footnotesize
\centering
\begin{tabular}{L{0.5cm}L{3.6cm}L{10.6cm}}
\toprule
\# & Check & Result \\
\midrule
1 & Length confound (original's Section F) & Length-matched test subsample (1,545 posts, median words human 1,004 vs AI 1,081): headline 96.9 vs 97.0 unmatched - unchanged, mirroring the original (93.2 to 93.2). Rarity d on the matched subset 1.83, corr(words, rarity) = 0.08. Per-model matched rarity means preserve the full-corpus ordering (deepseek 0.55 remains closest to humans) \\
2 & Publication-year check & Predicting a human post's publication period from structural features performs at chance (macro-F1 49.1, grouped CV; 187 vs 1,622 posts, where always guessing the larger class scores 47.3). Removing the most year-associated features leaves the headline flat (96.7-97.0). Top-25 year/authorship feature overlap 3/25 - publication year is not the signal \\
3 & Template-vs-direct discovery (the original's template-necessity ablation) & Same candidate yield (raw-on-text 354 vs template-run mean 356.7, all 11 dimensions) but only \textasciitilde 23\% bidirectional semantic overlap at 0.85 embedding similarity - the discovery pathway substantially shapes which features are found (original: 6 of top-20 overlap). The raw feature set was not scored at scale (declared) \\
4 & Deduplication cutoff sweep & Cutoffs 0.70-0.95 published: 200/233/251/266/277/282 features. Silhouette monotonically decreasing (0.182 at 0.70 to 0.065 at 0.85), so silhouette alone would prefer lower cutoffs - divergence disclosed. 0.85 retained for parity (merge 5.7\% vs their 25.5\%), sensitivity low \\
5 & Feature-discovery-set domain overlap & 11 of the 82 feature-discovery-set domains land in the test split. Sensitivity excluding them: 96.5 \\
6 & Memorization (original's Section E, both rules, exact implementations) & Exact 13-gram rule: 0.19\% of the 10,750 human-mirror pairs (the 2,150 classification prompts x 5 mirrors) share at least one 13-gram (shuffled-human control: 0.0\%, original: 0.70\% flagged). Near-verbatim rule (8-gram coverage >= 5\%, >= 4 distinct 8-grams, longest common span >= 30 tokens): 0 pairs. Excluding all 16 affected prompts: headline 97.2 (delta +0.3 points, original's filtered deltas <= +0.3) \\
7 & Vertical heterogeneity (Kruskal-Wallis, original's topic robustness) & H = 9.047, p = 0.107 over the 6 verticals with >= 20 test posts - not significant, consistent with the original's null (H = 4.69, p = 0.46). Per-vertical macro-F1 ranges from 94.8 (services) to 100 (health). Caveats: grouping uses >= 20 test posts where the original used >= 20 prompts (the thinnest vertical rests on 5 prompts), and per-post correctness observations are domain-clustered, an independence violation the original's version shares \\
8 & Learning curve & Train+val fractions 25/50/75/100\%: 95.4 / 96.4 / 96.4 / 97.0 (3 seeds each) \\
9 & Format sensitivity & The Section 4.5 rescoring repeated on the test split (250 AI posts; 235 human posts) flags 9 features, all among the 11 excluded. The structural classifier still detects 249 and 247 of 250 AI posts (markdown stripped; title also removed) and calls 15 of 235 human posts AI with the title added (18 as scored) \\
\bottomrule
\end{tabular}
\caption{Robustness checks, full results.}
\label{tab:robustnessfull}
\end{table*}

\section{Instrument validity, full detail}
\label{app:h}

This appendix carries the style-boundary audit behind Section 7.

\textbf{Style-boundary audit.} On 40 sampled features, human agreement with the model's binary style/non-style call: annotator A 90.0\% (kappa 0.765), annotator B 97.5\% (0.936), against a 0.75 bar fixed before the audit. Human-human agreement is 92.5\% (0.827). The strict style boundary (GPT-5.4, 3 runs, 0.989 inter-run agreement, 34 exclusions all inside the Writing-Style dimension) is endorsed by the human annotators.

\section{Original released-code defects and fixes}
\label{app:i}

This appendix records the five defects we found in the original's released code. Each is fixed in our fork toward the paper's \emph{stated} method, which makes the replication more faithful, not less. The unmodified originals and the full defect record are preserved in the release for side-by-side verification.

\begin{table*}[t]
\footnotesize
\centering
\begin{tabular}{L{0.5cm}L{3.8cm}L{5.2cm}L{4.9cm}}
\toprule
\# & Defect in released code & Symptom & Our fix \\
\midrule
B1 & Stage config keys passed twice to the provider layer & Every stage runner crashes on launch & De-duplicated the keyword pass-through \\
B2 & Stage 5 only recognizes a human\_story source column & All human posts silently dropped: the feature matrix would contain zero human rows & Source-column handling accepts the corpus's human and mirror columns \\
B3 & JSON-generation path ignores the thinking configuration & The paper states stage 5 runs ``minimal thinking'', but the released code sets no thinking budget and runs full reasoning (83x slower per dimension call) & Thinking budget set as a stage parameter, matching the paper's stated method \\
B4 & Shipped stage-3 prompt does not implement the paper's stated method & The prompt asks for a two-template quality comparison and contains \{group\_a\_json\}/\{group\_b\_json\} placeholders the code never fills, while the paper describes all six templates presented together with cross-source divergence mining & Prompt rewritten to the paper's specification (all templates together, structured per-source notes, divergences, executive summary). The shipped prompt and the first defective run are preserved \\
B5 & Released encoding deviates from the paper's stated scheme (deviation D9) & Ordinal and nominal features not encoded as the paper describes & Ordinal encoded by taxonomy position, nominal one-hot, per the paper's text \\
\bottomrule
\end{tabular}
\caption{Released-code defects and fixes.}
\label{tab:defects}
\end{table*}

Beyond these fixes and authentication/plumbing changes (provider base URLs and API-key paths, a 391-line fork patch over 7 files), no feature, encoding, classifier, or analysis code was modified.

\section{Rewording test detail}
\label{app:j}

This appendix carries the rewriting protocol, the verification gates, and the per-model detail behind Sections 4.8 and 5.4.

\textbf{Prompt design.} The rewriting prompt follows the LAMP taxonomy of \citet{chakrabarty2025lamp}: the seven categories of AI-writing artifacts their study identified with professional editors, together with their editing instructions. The prompt is instruction-only (deviation D15). A content-preservation clause instructs the model to keep every claim, fact, and link and to reword freely within them (deviation D16). The full prompt text is in the release.

\begin{table*}[t]
\small
\centering
\setlength{\tabcolsep}{4pt}
\begin{tabular}{L{2.4cm}L{2.1cm}L{2.1cm}}
\toprule
AI model & Surviving 13-gram share & Attacked structural macro-F1 \\
\midrule
gpt-5.4 & 0.175 & 96.5 \\
claude-sonnet-4.6 & 0.535 & 95.3 \\
gemini-3-flash & 0.003 & 95.9 \\
deepseek-v3.2 & 0.489 & 95.9 \\
kimi-k2.5 & 0.141 & 96.0 \\
\bottomrule
\end{tabular}
\caption{Per-model attack magnitude and attacked structural detection. The surviving 13-gram share is the fraction of a post's 13-word sequences that still appear verbatim after rewriting (mean 0.269 across models, i.e. 73\% of 13-word sequences replaced on average).}
\label{tab:attack}
\end{table*}

\textbf{Verification gates.} All gates were fixed before generation, and all passed. Length drift between a rewritten post and its source was bounded to the ratio interval [0.6, 1.4], with 0 violations (mean ratio 0.922, minimum 0.607). Refusals: 0 of 1,450. Trivial-copy flags: 0 after reruns. For claim preservation, we judged all 1,450 original-rewritten pairs in full: 90.3\% preserved every claim before quality control, and after a two-pass regeneration loop (192 regenerations) the full census reads 98.2\%, with 26 single-item residuals disclosed and left in.

\textbf{Rescoring and evaluation.} The rewritten posts are scored by the identical frozen stage-5 scorer and prompts (byte-identity asserted), covering all 15,950 answers (1,450 posts x 11 dimensions) with 0 unresolved failures and a rate of answers outside the predefined options of 0.082\% on average (0.089\% in our main scoring run). The encoding is identical, the classifiers are deterministic refits of the final models, and nothing is retrained on rewritten text. Of the 1,450 reworded AI posts, the structural classifier misclassifies 18 as human-written (9 before rewording), the style-only classifier 19 (8 before), and the all-features classifier 10 (3 before). Of these, 13, 16, and 8 posts respectively were classified correctly before rewording. Table J1 gives the attack magnitude and the attacked structural score per model.

\end{document}